\documentclass[runningheads,envcountsame,a4paper]{llncs}

\usepackage{microtype}
\usepackage{url}
\usepackage{graphicx}
\usepackage{amsfonts}
\usepackage{amsmath}
\usepackage{multirow}
\usepackage[linesnumbered,ruled]{algorithm2e}
\usepackage{pgfplots}
\usepackage{pgfplotstable} 
\pgfplotsset{compat=newest}

\begin{document}

\title{A Self-Attention Network based Node Embedding Model}

\author{Dai Quoc Nguyen${}^{1}$ \and Tu Dinh Nguyen${}^{2}$ \and Dinh Phung${}^{1}$}
%
%
\institute{${}^{1}$Monash University, Australia\\ \email{$\{$dai.nguyen,dinh.phung$\}$@monash.edu}\\
${}^{2}$\email{nguyendinhtu@gmail.com}}

\tocauthor{Dai Quoc Nguyen (Monash University),
Tu Dinh Nguyen (nguyendinhtu@gmail.com),
Dinh Phung (Monash University),
}
\toctitle{A Self-Attention Network based Node Embedding Model}

\maketitle \setcounter{footnote}{0}

\begin{abstract}

Despite several signs of progress have been made recently, limited research has been conducted for an inductive setting where embeddings are required for newly unseen nodes -- a setting encountered commonly in practical applications of deep learning for graph networks. This significantly affects the performances of downstream tasks such as node classification, link prediction or community extraction. To this end, we propose SANNE -- a novel unsupervised embedding model -- whose central idea is to employ a transformer self-attention network to iteratively aggregate vector representations of nodes in random walks. Our SANNE aims to produce plausible embeddings not only for present nodes, but also for newly unseen nodes. Experimental results show that the proposed SANNE obtains state-of-the-art results for the node classification task on well-known benchmark datasets. 

\end{abstract}

\section{Introduction}

Graph-structured data appears in plenty of fields in our real-world from social networks, citation networks, knowledge graphs and recommender systems to telecommunication networks, biological networks~\cite{hamilton2017representation,battaglia2018relational,cai2018comprehensive,Chen180802590}.
In graph-structured data, nodes represent individual entities, and edges represent relationships and interactions among those entities.
For example, in citation networks, each document is treated as a node, and a citation link between two documents is treated as an edge. 

Learning node embeddings is one of the most important and active research topics in representation learning for graph-structured data.
There have been many models proposed to embed each node into a continuous vector as summarized in~\cite{zhang2020network}.
These vectors can be further used in downstream tasks such as node classification, i.e., using learned node embeddings to train a classifier to predict node labels.
Existing models mainly focus on the \textit{transductive} setting where a model is trained using the entire input graph, i.e., the model requires all nodes with a fixed graph structure during training and 
lacks the flexibility in inferring embeddings for unseen/new nodes, e.g., DeepWalk~\cite{Perozzi:2014}, LINE~\cite{Tang:2015}, Node2Vec~\cite{Grover:2016}, SDNE~\cite{wang2016structural} and GCN~\cite{kipf2017semi}.
By contrast, a more important setup, but less mentioned, is the \textit{inductive} setting wherein only a part of the input graph is used to train the model, and then the learned model is used to infer embeddings for new nodes~\cite{Yang:2016planetoid}. Several attempts have additionally been made for the inductive settings such as EP-B~\cite{duran2017learning}, GraphSAGE~\cite{hamilton2017inductive} and GAT~\cite{velickovic2018graph}.
Working on the {inductive} setting is particularly more difficult than that on the {transductive} setting due to lacking the ability to generalize to the graph structure for new nodes.

One of the most convenient ways to learn node embeddings is to adopt the idea of a word embedding model by viewing each node as a word and each graph as a text collection of random walks to train a Word2Vec model~\cite{MikolovSCCD13nips}, e.g., DeepWalk, LINE and Node2Vec.
Although these Word2Vec-based approaches allow the current node to be directly connected with $k$-hops neighbors via random walks, they ignore feature information of nodes.
Besides, recent research has raised attention in developing graph neural networks (GNNs) for the node classification task, e.g., GNN-based models such as GCN, GraphSAGE and GAT.
These GNN-based models iteratively update vector representations of nodes over their $k$-hops neighbors using multiple layers stacked on top of each other.
Thus, it is difficult for the GNN-based models to infer plausible embeddings for new nodes when their $k$-hops neighbors are also unseen during training.

The transformer self-attention network~\cite{vaswani2017attention} has been shown to be very powerful in many NLP tasks such as machine translation and language modeling.
Inspired by this attention technique, we present SANNE -- an unsupervised learning model that adapts a transformer self-attention network to learn node embeddings. 
SANNE uses random walks (generated for every node) as inputs for a stack of attention layers.
Each attention layer consists of a self-attention sub-layer followed by a feed-forward sub-layer, wherein the self-attention sub-layer is constructed using query, key and value projection matrices to compute pairwise similarities among nodes.
Hence SANNE allows a current node at each time step to directly attend its $k$-hops neighbors in the input random walks.
SANNE then samples a set of ($1$-hop) neighbors for each node in the random walk and uses output vector representations from the last attention layer to infer embeddings for these neighbors.
As a consequence, our proposed SANNE produces the plausible node embeddings for both the transductive and inductive settings.

In short, our main contributions are as follows:

\begin{itemize}

\item Our SANNE induces a transformer self-attention network not only to work in the transductive setting advantageously, but also to infer the plausible embeddings for new nodes in the inductive setting effectively.

\item The experimental results show that our unsupervised SANNE obtains better results than up-to-date unsupervised and supervised embedding models on three benchmark datasets \textsc{Cora}, \textsc{Citeseer} and \textsc{Pubmed} for the transductive and inductive settings. 
In particular, SANNE achieves relative error reductions of more than 14\% over GCN and GAT in the inductive setting.

\end{itemize}

\section{Related work}

DeepWalk~\cite{Perozzi:2014} generates unbiased random walks starting from each node, considers each random walk as a sequence of nodes, and employs Word2Vec \cite{MikolovSCCD13nips} to learn node embeddings.
Node2Vec~\cite{Grover:2016} extends DeepWalk by introducing a biased random walk strategy that explores diverse neighborhoods and balances between exploration and exploitation from a given node. 
LINE~\cite{Tang:2015} closely follows Word2Vec, but introduces node importance, for which each node has a different weight to each of its neighbors, wherein weights can be pre-defined through algorithms such as PageRank~\cite{page1999pagerank}.
DDRW~\cite{LiJuzheng2016} jointly trains a DeepWalk model with a Support Vector Classification~\cite{Fan:2008} in a supervised manner.

SDNE~\cite{wang2016structural}, an autoencoder-based supervised model, is proposed to preserve both local and global graph structures.
EP-B~\cite{duran2017learning} is introduced to explore the embeddings of node attributes (such as words) with node neighborhoods to infer the embeddings of unseen nodes.
Graph Convolutional Network (GCN)~\cite{kipf2017semi}, a semi-supervised model, utilizes a variant of convolutional neural networks (CNNs) which makes use of layer-wise propagation to aggregate node features (such as profile information and text attributes) from the neighbors of a given node.
GraphSAGE~\cite{hamilton2017inductive} extends GCN in using node features and neighborhood structures to generalize to unseen nodes. 
Another extension of GCN is Graph Attention Network (GAT)~\cite{velickovic2018graph} that uses a similar idea with LINE~\cite{Tang:2015} in assigning different weights to different neighbors of a given node, but learns these weights by exploring an attention mechanism technique~\cite{bahdanau2014neural}.
These GNN-based approaches construct multiple layers stacked on top of each other to indirectly attend $k$-hops neighbors; thus, it is not straightforward for these approaches to infer the plausible embeddings for new nodes especially when their neighbors are also not present during training.

\section{The proposed SANNE}
\label{sec:ourmodel}

\begin{figure}[ht]
\centering
\includegraphics[width=0.75\textwidth]{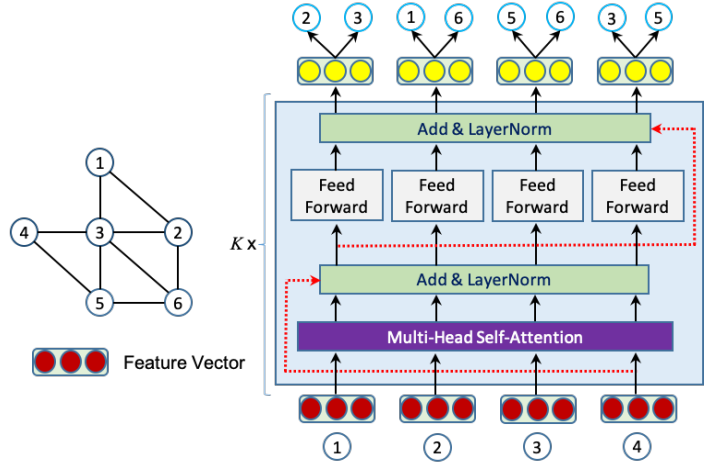}
\caption{Illustration of our SANNE learning process with $d = 3$, $N = 4$ and $M = 2$.}
\label{fig:SANNE}
\end{figure}

Let us define a graph as $\mathcal{G} = (\mathcal{V}, \mathcal{E})$, in which $\mathcal{V}$ is a set of nodes and $\mathcal{E}$ is a set of edges, i.e., $\mathcal{E} \subseteq \{(\mathsf{u},\mathsf{v}) | \mathsf{u}, \mathsf{v} \in \mathcal{V}\}$.
Each node $\mathsf{v} \in \mathcal{V}$ is associated with a feature vector $\boldsymbol{\mathsf{x}}_{\mathsf{v}} \in \mathbb{R}^{d}$ representing node features.
In this section, we detail the learning process of our proposed SANNE to learn a node embedding $\boldsymbol{\mathsf{o}}_\mathsf{v}$ for each node $\mathsf{v} \in \mathcal{V}$.
We then use the learned node embeddings to classify nodes into classes.

\textbf{SANNE architecture.} Particularly, we follow DeepWalk~\cite{Perozzi:2014} to uniformly sample random walks of length $N$ for every node in $\mathcal{V}$.
For example, Figure \ref{fig:SANNE} shows a graph consisting of 6 nodes where we generate a random walk of length $N = 4$ for node $1$, e.g., $\{1, 2, 3, 4\}$; and then this random walk is used as an input for the SANNE learning process.

Given a random walk \textit{w} of $N$ nodes $\{\mathsf{v}_{\textit{w},i}\}_{i=1}^N$, we obtain an input sequence of vector representations $\{\boldsymbol{\mathsf{u}}^{0}_{\mathsf{v}_{\textit{w},i}}\}_{i=1}^N$: $\boldsymbol{\mathsf{u}}^{0}_{\mathsf{v}_{\textit{w},i}} = \boldsymbol{\mathsf{x}}_{\mathsf{v}_{\textit{w},i}}$.
We construct a stack of $K$ attention layers~\cite{vaswani2017attention}, in which each of them has the same structure consisting of a multi-head self-attention sub-layer followed by a feed-forward sub-layer together with additionally using a residual connection~\cite{he2016deep} and layer normalization~\cite{ba2016layer} around each of these sub-layers.
At the $k$-th attention layer, we take an input sequence $\{\boldsymbol{\mathsf{u}}^{(k-1)}_{\mathsf{v}_{\textit{w},i}}\}_{i=1}^N$  
and produce an output sequence $\{\boldsymbol{\mathsf{u}}^{(k)}_{\mathsf{v}_{\textit{w},1}}\}_{i=1}^N$, $\boldsymbol{\mathsf{u}}^{(k)}_{\mathsf{v}_{\textit{w},1}} \in \mathbb{R}^{d}$ as:
\begin{eqnarray}
\boldsymbol{\mathsf{u}}^{(k)}_{\mathsf{v}_{\textit{w},i}} &=& \textsc{Walk-Transformer}\left(\boldsymbol{\mathsf{u}}^{(k-1)}_{\mathsf{v}_{\textit{w},i}}\right) \nonumber\\
\mathsf{In \ particular,} \ \ \ \boldsymbol{\mathsf{u}}^{(k)}_{\mathsf{v}_{\textit{w},i}} &=& \textsc{LayerNorm}\left(\boldsymbol{\mathsf{y}}^{(k)}_{\mathsf{v}_{\textit{w},i}} + \textsc{FF}\left(\boldsymbol{\mathsf{y}}^{(k)}_{\mathsf{v}_{\textit{w},i}}\right)\right) \nonumber\\
\mathsf{with} \ \ \ \boldsymbol{\mathsf{y}}^{(k)}_{\mathsf{v}_{\textit{w},i}} &=& \textsc{LayerNorm}\left(\boldsymbol{\mathsf{u}}^{(k-1)}_{\mathsf{v}_{\textit{w},i}} + \textsc{Att}\left(\boldsymbol{\mathsf{u}}^{(k-1)}_{\mathsf{v}_{\textit{w},i}}\right)\right)\nonumber
\end{eqnarray}
\noindent where $\textsc{FF}(.)$ and $\textsc{Att}(.)$ denote a two-layer feed-forward network and a multi-head self-attention network respectively:
\begin{eqnarray} 
\textsc{FF}\left(\boldsymbol{\mathsf{y}}^{(k)}_{\mathsf{v}_{\textit{w},i}}\right) = \textbf{W}^{(k)}_2\mathsf{ReLU}\left(\textbf{W}^{(k)}_1\boldsymbol{\mathsf{y}}^{(k)}_{\mathsf{v}_{\textit{w},i}} + \boldsymbol{\mathsf{b}}^{(k)}_1\right) + \boldsymbol{\mathsf{b}}^{(k)}_2 \nonumber
\end{eqnarray}
\noindent where $\textbf{W}^{(k)}_1$ and $\textbf{W}^{(k)}_2$ are weight matrices, and $\boldsymbol{\mathsf{b}}^{(k)}_1$ and $\boldsymbol{\mathsf{b}}^{(k)}_2$ are bias parameters. And:
\begin{eqnarray} 
\textsc{Att}\left(\boldsymbol{\mathsf{u}}^{(k-1)}_{\mathsf{v}_{\textit{w},i}}\right) = \textbf{W}^{(k)}\left[\boldsymbol{\mathsf{h}}^{(k),1}_{\mathsf{v}_{\textit{w},i}} ;  \boldsymbol{\mathsf{h}}^{(k),2}_{\mathsf{v}_{\textit{w},i}} ; ... ; \boldsymbol{\mathsf{h}}^{(k),H}_{\mathsf{v}_{\textit{w},i}}\right] \nonumber
\end{eqnarray}
\noindent where $\textbf{W}^{(k)} \in \mathbb{R}^{d\times Hs}$ is a weight matrix, $H$ is the number of attention heads, and $\left[ ; \right]$ denotes a vector concatenation. 
Regarding the $h$-th attention head, $\boldsymbol{\mathsf{h}}^{(k),h}_{\mathsf{v}_{\textit{w},i}} \in \mathbb{R}^{s}$ is calculated by a weighted sum as:
\begin{eqnarray}
\boldsymbol{\mathsf{h}}^{(k),h}_{\mathsf{v}_{\textit{w},i}} = \sum_{j=1}^N\alpha^{(k)}_{i,j,h}\left(\textbf{W}^{(k),V}_h\boldsymbol{\mathsf{u}}^{(k-1)}_{\mathsf{v}_{\textit{w},j}}\right) \nonumber
\end{eqnarray}
\noindent where $\textbf{W}^{(k),V}_h \in \mathbb{R}^{s\times d}$ is a value projection matrix, and $\alpha^{(k)}_{i,j,h}$ is an attention weight. $\alpha^{(k)}_{i,j,h}$ is computed using the $\mathsf{softmax}$  function over scaled dot products between $i$-th and $j$-th nodes in the walk $w$ as:
\begin{eqnarray}
\alpha^{(k)}_{i,j,h} = \mathsf{softmax}\left(\frac{\left(\textbf{W}^{(k),Q}_h\boldsymbol{\mathsf{u}}^{(k-1)}_{\mathsf{v}_{\textit{w},i}}\right)^\mathsf{T}\left(\textbf{W}^{(k),K}_h\boldsymbol{\mathsf{u}}^{(k-1)}_{\mathsf{v}_{\textit{w},j}}\right)}{\sqrt{k}}\right) \nonumber
\end{eqnarray}
\noindent where $\textbf{W}^{(k),Q}_h$ and $\textbf{W}^{(k),K}_h \in \mathbb{R}^{s\times d}$ are query and key  projection matrices \mbox{respectively}.

We randomly sample a fixed-size set of $M$ neighbors for each node in the random walk \textit{w}.
We then use the output vector representations $\boldsymbol{\mathsf{u}}^{(K)}_{\mathsf{v}_{\textit{w},i}}$ from the $K$-th last layer to infer embeddings $\boldsymbol{\mathsf{o}}$ for sampled neighbors of $\mathsf{v}_{\textit{w},i}$.
Figure \ref{fig:SANNE} illustrates our proposed SANNE where we set the length $N$ of random walks to 4, the dimension size $d$ of feature vectors to 3, and the number $M$ of sampling neighbors to 2.
We also sample different sets of neighbors for the same input node at each training step.

\begin{algorithm}[h]
\DontPrintSemicolon
\SetAlgoVlined

\textbf{Input}: A network graph $\mathcal{G} = (\mathcal{V}, \mathcal{E})$.

\For{$\mathsf{v} \in \mathcal{V}$}{
	\textsc{Sample} $T$ random walks of length $N$ rooted by $\mathsf{v}$.
}

\For{each $\mathsf{random\ walk}$ \textit{w}}{
    
    \For{k = 1, 2, ..., K}{

$\forall \mathsf{v} \in \textit{w}$

$\boldsymbol{\mathsf{y}}^{(k)}_{\mathsf{v}} \leftarrow \textsc{LayerNorm}\left(\boldsymbol{\mathsf{u}}^{(k-1)}_{\mathsf{v}} + \textsc{Att}\left(\boldsymbol{\mathsf{u}}^{(k-1)}_{\mathsf{v}}\right)\right)$


$\boldsymbol{\mathsf{u}}^{(k)}_{\mathsf{v}} \leftarrow \textsc{LayerNorm}\left(\boldsymbol{\mathsf{y}}^{(k)}_{\mathsf{v}} + \textsc{FF}\left(\boldsymbol{\mathsf{y}}^{(k)}_{\mathsf{v}}\right)\right)$
    }
    
    \For{$\mathsf{v} \in \textit{w}$}{
\textsc{Sample} a set $\mathsf{C}_\mathsf{v}$ of $M$ neighbors of node $\mathsf{v}$.

$\boldsymbol{\mathsf{o}}_{\mathsf{v'}} \leftarrow \boldsymbol{\mathsf{u}}^{(K)}_{\mathsf{v}}, \forall \mathsf{v}' \in \mathsf{C}_\mathsf{v}$
    }
}
\caption{The SANNE learning process.}
\label{alg:SANNE}
\end{algorithm}

\textbf{Training SANNE:}
We learn our model's parameters including the weight matrices and node embeddings by minimizing the sampled softmax loss function~\cite{Jean2015} applied to the random walk \textit{w} as:
\begin{equation}
\mathcal{L}_{\mathsf{SANNE}}\left(\textit{w}\right) = - \sum_{i=1}^N\sum_{\mathsf{v}' \in \mathsf{C}_{\mathsf{v}_{\textit{w},i}}}\log \frac{\exp(\boldsymbol{\mathsf{o}}_\mathsf{v'}^\mathsf{T} \boldsymbol{\mathsf{u}}^{(K)}_{\mathsf{v}_{\textit{w},i}})}{\sum_{\mathsf{u} \in \mathcal{V'}} \exp(\boldsymbol{\mathsf{o}}_\mathsf{u}^\mathsf{T} \boldsymbol{\mathsf{u}}^{(K)}_{\mathsf{v}_{\textit{w},i}})} \nonumber
\label{equa:SANNE}
\end{equation}
\noindent where $\mathsf{C}_\mathsf{v}$ is the fixed-size set of $M$ neighbors randomly sampled for node $\mathsf{v}$, $\mathcal{V'}$ is a subset sampled from $\mathcal{V}$, and $\boldsymbol{\mathsf{o}}_\mathsf{v} \in \mathbb{R}^{d}$ is the node embedding of node $\mathsf{v}, \forall \mathsf{v} \in \mathcal{V}$.
Node embeddings $\boldsymbol{\mathsf{o}}_\mathsf{v}$ are learned implicitly as model parameters.

We briefly describe the learning process of our proposed SANNE model in Algorithm \ref{alg:SANNE}. Here, the learned node embeddings  $\boldsymbol{\mathsf{o}}_\mathsf{v}$ are used as the final representations of nodes $\mathsf{v}\in\mathcal{V}$.
We explicitly aggregate node representations from both left-to-right and right-to-left sides in the walk for each node in predicting its neighbors.
This allows SANNE to infer the plausible node embeddings even in the inductive setting.

\begin{algorithm}[ht]
{
\DontPrintSemicolon
\SetAlgoVlined
\textbf{Input}: A network graph $\mathcal{G} = (\mathcal{V}, \mathcal{E})$, a trained model SANNE$_{trained}$ for $\mathcal{G}$, a set $\mathcal{V}_{new}$ of new nodes.

\For{$v \in \mathcal{V}_{new}$}{
	\textsc{Sample} $Z$ random walks \{\textit{w}$_i$\}$_{i=1}^Z$ of length $N$ rooted by $v$.
    
\For{$i \in \{1,\ 2,\ ...,\ Z\}$}{
    $\boldsymbol{\mathsf{u}}^{(K)}_{v,i} \leftarrow$ SANNE$_{trained}\left(\textit{w}_i\right)[0]$
}

$\boldsymbol{\mathsf{o}}_v \leftarrow \textsc{Average}\left(\{\boldsymbol{\mathsf{u}}^{(K)}_{v,i}\}_{i=1}^Z\right)$
}
}
\caption{The embedding inference for new nodes.}
\label{alg:InferInd}
\end{algorithm}

\textbf{Inferring embeddings for new nodes in the inductive setting:}
After training our SANNE on a given graph, we show in Algorithm \ref{alg:InferInd} our method to infer an embedding for a new node $v$ adding to this given graph. 
We randomly sample $Z$ random walks of length $N$ starting from $v$. 
We use each of these walks as an input for our trained model and then collect the first vector representation (at the index 0 corresponding to node $v$) from the output sequence at the $K$-th last layer.
Thus, we obtain $Z$ vectors and then average them into a final embedding for the new node $v$.

\section{Experiments}
Our SANNE is evaluated for the node classification task as follows:
(i) We train our model to obtain node embeddings.
(ii) We use these node embeddings to learn a logistic regression classifier to predict node labels.
(iii) We evaluate the classification performance on benchmark datasets and then analyze the effects of hyper-parameters.

\subsection{Datasets and data splits}

\begin{table}[!ht]
\centering
\caption{Statistics of the experimental datasets.
$|$Vocab.$|$ denotes the vocabulary size. Avg.W denotes the average number of words per node.
}
\def\arraystretch{1.1}
\setlength{\tabcolsep}{0.75em}
\begin{tabular}{l|llclc}
\hline
\bf Dataset &  \bf{$|\mathcal{V}|$} & \bf{$|\mathcal{E}|$} & {\#Classes} & $|$Vocab.$|$ & Avg.W\\
\hline
\textsc{Cora} & 2,708 & 5,429 & 7 & 1,433 & 18\\
\textsc{Citeseer} & 3,327 & 4,732 & 6 & 3,703 & 31\\
\textsc{Pubmed} & 19,717 & 44,338 & 3 & 500 & 50\\
\hline
\end{tabular}
\label{tab:graphdatasets}
\end{table}

\subsubsection{Datasets}

We use three well-known benchmark datasets \textsc{Cora, Citeseer}~\cite{sen2008collective} and \textsc{Pubmed}~\cite{namata:mlg12} which are citation networks.
For each dataset, each node represents a document, and each edge represents a citation link between two documents. 
Each node is assigned a class label representing the main topic of the document. 
Besides, each node is also associated with a feature vector of a bag-of-words.
Table \ref{tab:graphdatasets} reports the statistics of these three datasets.

\subsubsection{Data splits} 
We follow the same settings used in~\cite{duran2017learning} for a fair comparison. 
For each dataset, we uniformly sample 20 random nodes for each class as training data, 1000 different random nodes as a validation set, and 1000 different random nodes as a test set. 
We repeat 10 times to have 10 training sets, 10 validation sets, and 10 test sets respectively, and finally report the mean and standard deviation of the accuracy results over 10 data splits.

\subsection{Training protocol}
\label{subsec:training}

\subsubsection{Feature vectors initialized by Doc2Vec} 

For each dataset, each node represents a document associated with an existing feature vector of a bag-of-words.
Thus, we train a PV-DBOW Doc2Vec model~\cite{le:2014} to produce new 128-dimensional embeddings $\boldsymbol{\mathsf{x}}_{\mathsf{v}}$ which are considered as new feature vectors for nodes $\mathsf{v}$.
Using this initialization is convenient and efficient for our proposed SANNE compared to using the feature vectors of bag-of-words.

\subsubsection{Positional embeddings} 

We hypothesize that the relative positions among nodes in the random walks are useful to provide meaningful information about the graph structure.
Hence we add to each position $i$ in the random walks a pre-defined positional embedding $\boldsymbol{\mathsf{t}}_i \in \mathbb{R}^{d}, i \in \{1, 2, ..., N\}$ using the sinusoidal functions~\cite{vaswani2017attention}, so that we can use
$\boldsymbol{\mathsf{u}}^{0}_{\mathsf{v}_{\textit{w},i}} = \boldsymbol{\mathsf{x}}_{\mathsf{v}_{\textit{w},i}} + \boldsymbol{\mathsf{t}}_i$ where $\mathsf{t}_{i,2j} = \sin(i/10000^{2j/d})$ and $\mathsf{t}_{i,2j+1} = \cos(i/10000^{2j/d})$.
From preliminary experiments, adding the positional embeddings produces better performances on \textsc{Cora} and \textsc{Pubmed}; thus, we keep to use the positional embeddings on these two datasets.

\subsubsection{Transductive setting} 
\textit{This setting is used in most of the existing approaches where we use the entire input graph, i.e., all nodes are present during training}.
We fix the dimension size $d$ of feature vectors and node embeddings to 128 ($d = 128$ with respect to the Doc2Vec-based new feature vectors), the batch size to 64, the number $M$ of sampling neighbors to 4 ($M = 4$) and the number of samples in the sampled loss function to 512 ($|\mathcal{V'}| = 512$).
We also sample $T$ random walks of a fixed length $N$ = 8 starting from each node, wherein $T$ is empirically varied in \{16, 32, 64, 128\}.
We vary the hidden size of the feed-forward sub-layers in \{1024, 2048\}, the number $K$ of attention layers in \{2, 4, 8\} and the number $H$ of attention heads in \{4, 8, 16\}. 
The dimension size $s$ of attention heads is set to satisfy that $Hs = d$.

\subsubsection{Inductive setting} We use the same inductive setting as used in~\cite{Yang:2016planetoid,duran2017learning}. \textit{Specifically, for each of 10 data splits, we first remove all 1000 nodes in the test set from the original graph before the training phase, so that these nodes are becoming unseen/new in the testing/evaluating phase}.
We then apply the standard training process on the resulting graph.
From preliminary experiments, we set the number $T$ of random walks sampled for each node on \textsc{Cora} and \textsc{Pubmed} to 128 ($T=128$), and on \textsc{Citeseer} to 16 ($T=16$).
Besides, we adopt the same value sets of other hyper-parameters for tuning as used in the transductive setting to train our SANNE in this inductive setting.
After training, we infer the embedding for each unseen/new node $\mathsf{v}$ in the test set as described in Algorithm \ref{alg:InferInd} with setting $Z=8$.

\subsubsection{Training SANNE to learn node embeddings}
For each of the 10 data splits, to learn our model parameters in the transductive and inductive settings, we use the Adam optimizer~\cite{kingma2014adam} to train our model and select the initial learning rate in $\{1e^{-5}, 5e^{-5}, 1e^{-4}\}$.
We run up to 50 epochs and evaluate the model for every epoch to choose the best model on the validation set.

\subsection{Evaluation protocol}
\label{subsec:eval}
We also follow the same setup used in~\cite{duran2017learning} for the node classification task. 
For each of the 10 data splits, we use the learned node embeddings as feature inputs to learn a L2-regularized logistic regression classifier~\cite{Fan:2008} on the training set.
We monitor the classification accuracy on the validation set for every training epoch, and take the model that produces the highest accuracy on the validation set to compute the accuracy on the test set. 
We finally report the mean and standard deviation of the accuracies across 10 test sets in the 10 data splits.

\textbf{Baseline models:} We compare our unsupervised SANNE with previous unsupervised models including DeepWalk (DW), Doc2Vec and EP-B; and previous supervised models consisting of Planetoid, GCN and GAT.
Moreover, as reported in~\cite{guo2018spine}, GraphSAGE obtained low accuracies on \textsc{Cora}, \textsc{Pubmed} and \textsc{Citeseer}, {thus we do not include GraphSAGE as a strong baseline}.

The results of DeepWalk (DW), DeepWalk+BoW (DW+BoW), Planetoid, GCN and EP-B are taken from~\cite{duran2017learning}.\footnote{As compared to our experimental results for Doc2Vec and GAT, showing the statistically significant differences for DeepWalk, Planetoid, GCN and EP-B against our SANNE in Table \ref{tab:expresults} is justifiable.} 
Note that DeepWalk+BoW denotes a concatenation between node embeddings learned by DeepWalk and the bag-of-words feature vectors.
Regarding the inductive setting for DeepWalk,~\cite{duran2017learning} computed the embeddings for new nodes by averaging the embeddings of their neighbors.
In addition, we provide our new results for Doc2Vec and GAT using our experimental setting.

\begin{table}[!ht]
\centering
\caption{Experimental results on the \textsc{Cora}, \textsc{Pubmed} and \textsc{Citeseer} {test} sets in the transductive and inductive settings across the 10 data splits. The best score is in {bold}, while the second-best score is in {underline}. 
``{Unsup}'' denotes a group of unsupervised models. ``{Sup}'' denotes a group of supervised models using node labels from the training set during training. ``{Semi}'' denotes a group of semi-supervised models also using node labels from the training set together with node feature vectors from the entire dataset during training.
$\ast$ denotes the statistically significant differences against our SANNE at $p < 0.05$ (using the two-tailed \textit{paired t-test}).
Numeric subscripts denote the relative error reductions over the baselines.
\textit{Note that the inductive setting~\cite{Yang:2016planetoid,duran2017learning} is used to evaluate the models when we do not access nodes in the test set during training.
This inductive setting was missed in the original GCN and GAT papers which relied on the semi-supervised training process} for \textsc{Cora}, \textsc{Pubmed} and \textsc{Citeseer}. 
Regarding the inductive setting on Cora and Citeseer, many neighbors of test nodes also belong to the test set, thus these neighbors are unseen during training and then become new nodes in the testing/evaluating phase.
}
\def\arraystretch{1.1}
\setlength{\tabcolsep}{0.75em}
\begin{tabular}{l|l|l|l|l}
\hline
\multicolumn{2}{c|}{\bf Transductive} &  \textbf{\textsc{Cora}} &  \textbf{\textsc{Pubmed}} &  \textbf{\textsc{Citeseer}}\\
\hline
\multirow{5}{*}{\rotatebox[origin=c]{90}{\textbf{Unsup}}} & DW~\cite{Perozzi:2014} & 71.11 $\pm$ 2.70$^\ast_{33.6}$ & 73.49 $\pm$ 3.00$^\ast_{23.3}$ & 47.60 $\pm$ 2.34$^\ast_{43.0}$\\
& DW+BoW & 76.15 $\pm$ 2.06$^\ast_{19.6}$ & 77.82 $\pm$ 2.19$^\ast_{8.3}$ & 61.87 $\pm$ 2.30$^\ast_{21.8}$\\
& Doc2Vec~\cite{le:2014} & 64.90 $\pm$ 3.07$^\ast_{45.4}$ & 76.12 $\pm$ 1.62$^\ast_{14.9}$ & 64.58 $\pm$ 1.84$^\ast_{15.8}$\\
& EP-B~\cite{duran2017learning} & \underline{78.05 $\pm$ 1.49}$^\ast_{12.6}$ & \underline{79.56 $\pm$ 2.10}$_{0.5}$ & \textbf{71.01 $\pm$ 1.35}$_{-2.9}$\\
& Our \textbf{SANNE} & \textbf{80.83 $\pm$ 1.94} & \textbf{79.67 $\pm$ 1.28}& \underline{70.18 $\pm$ 2.12}\\
\hline
\multirow{3}{*}{\rotatebox[origin=c]{90}{\textbf{Semi}}} & GCN~\cite{kipf2017semi} & 79.59 $\pm$ 2.02$_{6.1}$ & 77.32 $\pm$ 2.66$^\ast_{10.4}$ & 69.21 $\pm$ 1.25$_{3.1}$\\
& GAT~\cite{velickovic2018graph} & {81.72 $\pm$ 2.93$_{-4.8}$} & 79.56 $\pm$ 1.99$_{0.5}$ & 70.80 $\pm$ 0.92$_{-2.1}$\\
& Planetoid~\cite{Yang:2016planetoid} & 71.90 $\pm$ 5.33$^\ast_{31.7}$ & 74.49 $\pm$ 4.95$^\ast_{20.3}$ & 58.58 $\pm$ 6.35$^\ast_{28.0}$\\
\hline
\hline
\multicolumn{2}{c|}{\bf Inductive} &  \textbf{\textsc{Cora}} & \textbf{\textsc{Pubmed}} & \textbf{\textsc{Citeseer}} \\
\hline
\multirow{3}{*}{\rotatebox[origin=c]{90}{\textbf{Unsup}}} & DW+BoW  & 68.35 $\pm$ 1.70$^\ast_{26.5}$ & 74.87 $\pm$ 1.23$^\ast_{20.6}$ & 59.47 $\pm$ 2.48$^\ast_{23.1}$\\
& EP-B~\cite{duran2017learning} & \underline{73.09 $\pm$ 1.75}$^\ast_{13.6}$ & \underline{79.94 $\pm$ 2.30}$_{0.5}$ & \underline{68.61 $\pm$ 1.69}$_{0.7}$\\
& Our \textbf{SANNE} & \textbf{76.75 $\pm$ 2.45} & \textbf{80.04 $\pm$ 1.67} & \textbf{68.82 $\pm$ 3.21} \\
\hline
\multirow{3}{*}{\rotatebox[origin=c]{90}{\textbf{Sup}}} & GCN~\cite{kipf2017semi} & 67.76 $\pm$ 2.11$^\ast_{27.9}$ & 73.47 $\pm$ 2.48$^\ast_{24.8}$ & 63.40 $\pm$ 0.98$^\ast_{14.8}$\\
& GAT~\cite{velickovic2018graph} & {69.37 $\pm$ 3.81$^\ast_{24.1}$} & {71.29 $\pm$ 3.56$^\ast_{30.5}$} & 59.55 $\pm$ 4.21$^\ast_{22.9}$\\
& Planetoid~\cite{Yang:2016planetoid} & 64.80 $\pm$ 3.70$^\ast_{33.9}$ & {75.73 $\pm$ 4.21$^\ast_{17.8}$} & 61.97 $\pm$ 3.82$^\ast_{18.0}$\\
\hline
\end{tabular}
\label{tab:expresults}
\end{table}

\subsection{Main results}

Table \ref{tab:expresults} reports the experimental results in the transductive and inductive settings where the best scores are in {bold}, while the second-best in {underline}.
As discussed in~\cite{duran2017learning}, the experimental setup used for GCN and GAT~\cite{kipf2017semi,velickovic2018graph} is not fair enough to show the effectiveness of existing models when the models are evaluated {only using the fixed training, validation and test sets split by~\cite{Yang:2016planetoid}}, thus we do not rely on the GCN and GAT results reported in the original papers. Here, we do include the accuracy results of GCN and GAT using the same settings used in~\cite{duran2017learning}.

Regarding the transductive setting, SANNE obtains the highest scores on \textsc{Cora} and \textsc{Pubmed}, and the second-highest score on \textsc{Citeseer} {in the group of unsupervised models}.
In particular, SANNE works better than EP-B on \textsc{Cora}, while both models produce similar scores on \textsc{Pubmed}.
Besides, SANNE produces high competitive results compared to the up-to-date semi-supervised models GCN and GAT.
Especially, SANNE outperforms GCN with relative error reductions of 6.1\% and 10.4\% on \textsc{Cora} and \textsc{Pubmed} respectively.
Furthermore, it is noteworthy that there is {no} statistically significant difference between SANNE and GAT at $p < 0.05$ (using the two-tailed {paired t-test})on these datasets.

EP-B is more appropriate than other models for \textsc{Citeseer} in the transductive setting because (i) EP-B simultaneously learns word embeddings from the texts within nodes, which are then used to reconstruct the embeddings of nodes from their neighbors; (ii) \textsc{Citeseer} is quite sparse; thus word embeddings can be useful in learning the node embeddings.
But we emphasize that using a significant test, there is \textit{no} difference between EP-B and our proposed SANNE on \textsc{Citeseer}; hence the results are comparable.

More importantly, regarding the inductive setting, SANNE obtains the highest scores on three benchmark datasets, hence these show the effectiveness of SANNE in inferring the plausible embeddings for new nodes.
Especially, SANNE outperforms both GCN and GAT in this setting, e.g., SANNE achieves absolute improvements of 8.9\%, 6.6\% and 5.4\% over GCN, and 7.3\%, 8.7\% and 9.3\% over GAT, on \textsc{Cora}, \textsc{Pubmed} and \textsc{Citeseer}, respectively (with relative error reductions of more than 14\% over GCN and GAT).

\begin{figure}[!ht]
\centering
    \includegraphics[width=0.475\textwidth]{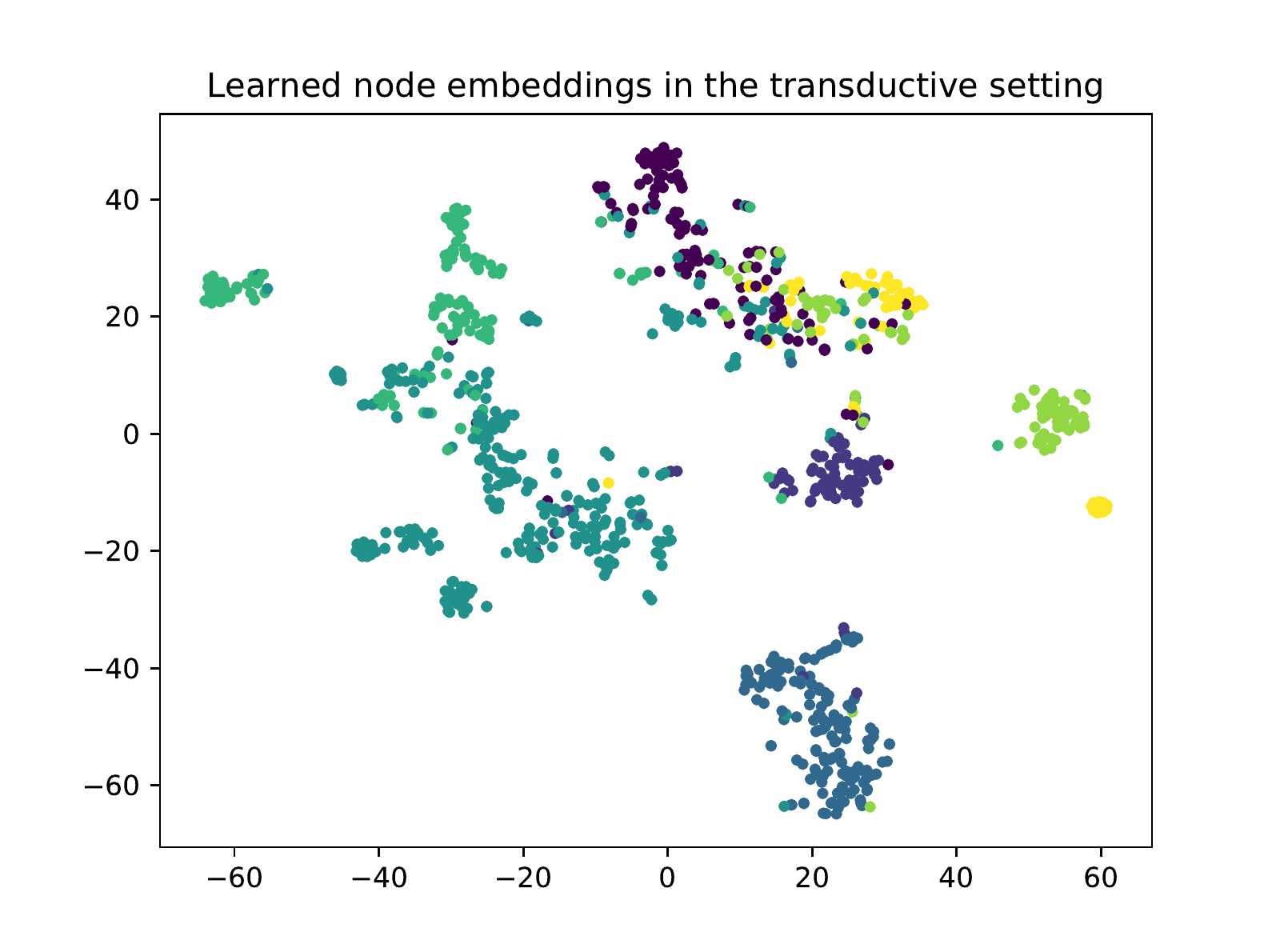}
    \includegraphics[width=0.475\textwidth]{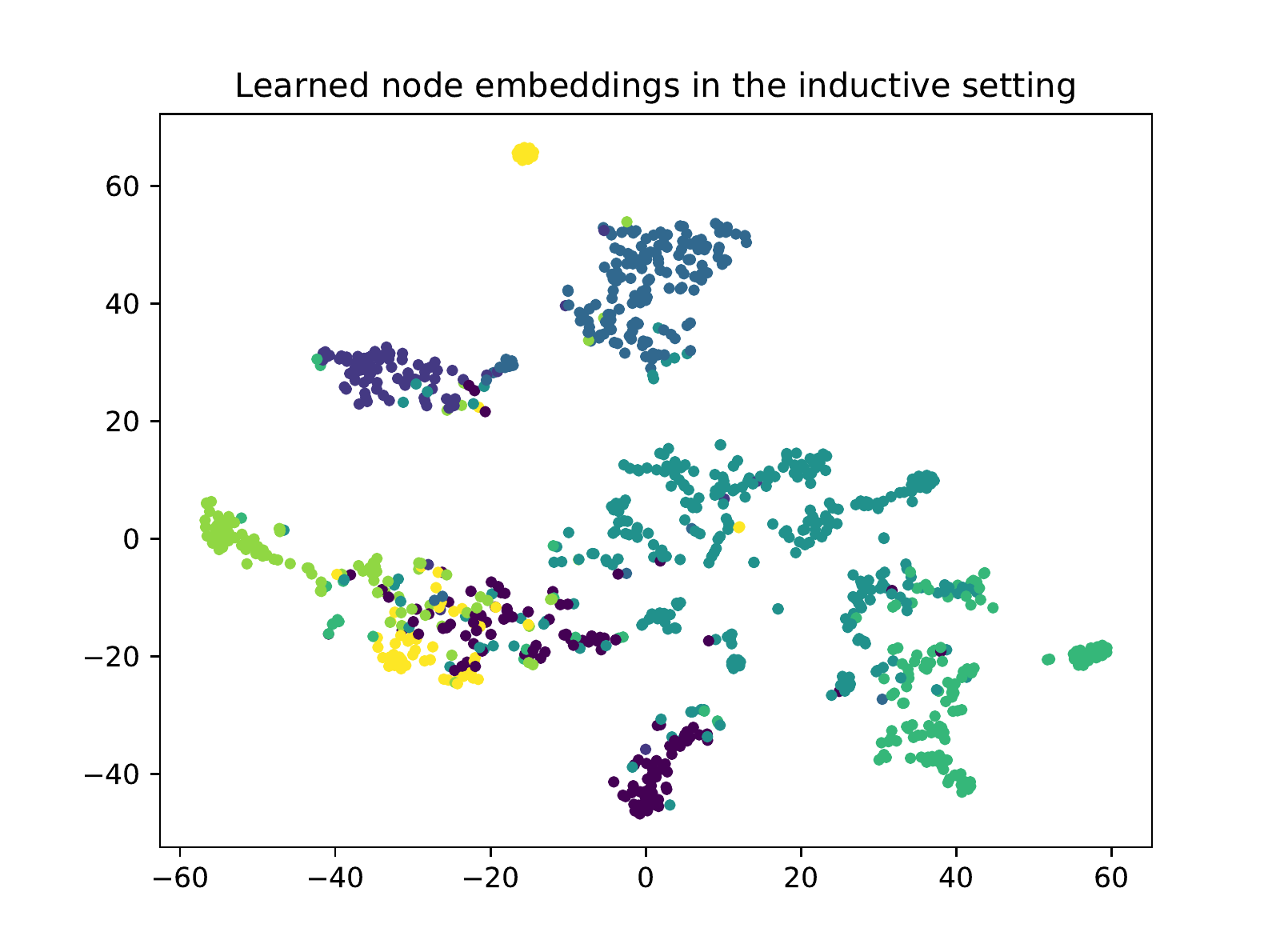}
\caption{A visualization of the learned node embeddings in the transductive and inductive settings for one data split on \textsc{Cora}.}

\label{fig:visualembeddings}
\end{figure}

Compared to the transductive setting, the inductive setting is particularly difficult due to requiring the ability to align newly observed nodes to the present nodes. 
As shown in Table \ref{tab:expresults}, there is a significant decrease for GCN and GAT from the transductive setting to the inductive setting on all three datasets, while by contrast, our SANNE produces reasonable accuracies for both settings. 
To qualitatively demonstrate this advantage of SANNE, we use t-SNE~\cite{maaten2008visualizing} to visualize the learned node embeddings on one data split of the \textsc{Cora} dataset in Figure \ref{fig:visualembeddings}. We see a similarity in the node embeddings (according to their labels) between two settings, verifying the plausibility of the node embeddings learned by our SANNE in the inductive setting.

\subsection{Ablation analysis}

\begin{table}[!ht]
\centering
\caption{Ablation results on the {validation} sets in the transductive setting. (i) Without using the feed-forward sub-layer: $\boldsymbol{\mathsf{u}}^{(k)}_{\mathsf{v}} = \textsc{LayerNorm}\left(\boldsymbol{\mathsf{u}}^{(k-1)}_{\mathsf{v}} + \textsc{Att}\left(\boldsymbol{\mathsf{u}}^{(k-1)}_{\mathsf{v}}\right)\right)$ (ii) Without using the multi-head self-attention sub-layer: $\boldsymbol{\mathsf{u}}^{(k)}_{\mathsf{v}} = \textsc{LayerNorm}\left(\boldsymbol{\mathsf{u}}^{(k-1)}_{\mathsf{v}} + \textsc{FF}\left(\boldsymbol{\mathsf{u}}^{(k-1)}_{\mathsf{v}}\right)\right)$. $\ast$ denotes the statistically significant differences at $p < 0.05$ (using the two-tailed {paired t-test}).
}
\def\arraystretch{1.25}
\setlength{\tabcolsep}{0.75em}
\begin{tabular}{l|l|l|l}
\hline
\bf Transductive &  \textbf{\textsc{Cora}} &  \textbf{\textsc{Pubmed}} &  \textbf{\textsc{Citeseer}}\\
\hline
Our \textbf{SANNE} & 81.32 $\pm$ 1.20 &  78.28 $\pm$ 1.24 & 70.77 $\pm$ 1.18\\
\hline
 \ \ \ \ \ (i) w/o FF & 80.77 $\pm$ 1.34 & 77.90 $\pm$ 1.76 & 70.36 $\pm$ 1.32 \\
 \ \ \ \ \ (ii) w/o ATT & 77.87 $\pm$ 1.09$^\ast$ & 74.52 $\pm$ 2.66$^\ast$ & 65.68 $\pm$ 1.31$^\ast$\\
\hline
\end{tabular}
\label{tab:expresultsDev}
\end{table}

\begin{figure}[!htb]
\centering
\centering
\resizebox{12cm}{!}{
\begin{tikzpicture}
\begin{axis}[
    ybar,
    enlarge x limits=0.25,
    legend style={at={(0.5,1)},
                anchor=north,legend columns=3},
    ylabel={Accuracy},
    xlabel={Transductive},
    symbolic x coords={$K$=2, $K$=4, $K$=8},
    xtick=data,
    ymin=68,ymax=85,
    nodes near coords={\scriptsize\pgfmathprintnumber\pgfplotspointmeta},
    ]
\addplot+[error bars/.cd, y dir=both,y explicit] coordinates {
($K$=2,81.16) +- (0.0, 1.11)
($K$=4,81.19) +- (0.0, 1.29) 
($K$=8,80.96) +- (0.0, 1.35)
};
\addplot+[error bars/.cd, y dir=both,y explicit] coordinates {
($K$=2,77.62) +- (0.0, 1.80) 
($K$=4,78.00) +- (0.0, 2.05)
($K$=8,78.11) +- (0.0, 1.86)
};
\addplot+[error bars/.cd, y dir=both,y explicit] coordinates {
($K$=2,70.13) +- (0.0, 1.44) 
($K$=4,70.41) +- (0.0, 1.18)
($K$=8,70.78) +- (0.0, 1.29)
};
\legend{\textsc{Cora},\textsc{Pubmed},\textsc{Citeseer}}
\end{axis}
\end{tikzpicture}

\begin{tikzpicture}
\begin{axis}[
    ybar,
    enlarge x limits=0.25,
    legend style={at={(0.5,1)},
                anchor=north,legend columns=3},
    ylabel={Accuracy},
    xlabel={Inductive},
    symbolic x coords={$K$=2, $K$=4, $K$=8},
    xtick=data,
    ymin=58,ymax=82,
    nodes near coords={\scriptsize\pgfmathprintnumber\pgfplotspointmeta},
    ]
\addplot+[error bars/.cd, y dir=both,y explicit] coordinates {
($K$=2,72.32) +- (0.0, 1.29)
($K$=4,72.10) +- (0.0, 1.32) 
($K$=8,72.08) +- (0.0, 1.14)
};
\addplot+[error bars/.cd, y dir=both,y explicit] coordinates {
($K$=2,76.32) +- (0.0, 2.00)
($K$=4,76.42) +- (0.0, 1.91) 
($K$=8,76.53) +- (0.0, 2.07)
};
\addplot+[error bars/.cd, y dir=both,y explicit] coordinates {
($K$=2,61.37) +- (0.0, 1.53)
($K$=4,61.81) +- (0.0, 1.52)
($K$=8,62.20) +- (0.0, 1.42)
};
\legend{\textsc{Cora},\textsc{Pubmed},\textsc{Citeseer}}
\end{axis}
\end{tikzpicture}
}
\\
\centering
\resizebox{12cm}{!}{
\begin{tikzpicture}
\begin{axis}[
    ybar,
    enlarge x limits=0.25,
    legend style={at={(0.5,1)},
                anchor=north,legend columns=3},
    ylabel={Accuracy},
    xlabel={Transductive},
    symbolic x coords={$H$=4, $H$=8, $H$=16},
    xtick=data,
    ymin=68,ymax=85,
    nodes near coords={\scriptsize\pgfmathprintnumber\pgfplotspointmeta},
    ]
\addplot+[error bars/.cd, y dir=both,y explicit] coordinates {
($H$=4,80.97) +- (0.0, 1.27)
($H$=8,81.25) +- (0.0, 1.20) 
($H$=16,81.1) +- (0.0, 1.24)
};
\addplot+[error bars/.cd, y dir=both,y explicit] coordinates {
($H$=4,78.05) +- (0.0, 2.15) 
($H$=8,78.02) +- (0.0, 1.76)
($H$=16,78.05) +- (0.0, 1.92)
};
\addplot+[error bars/.cd, y dir=both,y explicit] coordinates {
($H$=4,70.58) +- (0.0, 1.25) 
($H$=8,70.55) +- (0.0, 1.43)
($H$=16,70.60) +- (0.0, 1.12)
};
\legend{\textsc{Cora},\textsc{Pubmed},\textsc{Citeseer}}
\end{axis}
\end{tikzpicture}

\begin{tikzpicture}
\begin{axis}[
    ybar,
    enlarge x limits=0.25,
    legend style={at={(0.5,1)},
                anchor=north,legend columns=3},
    ylabel={Accuracy},
    xlabel={Inductive},
    symbolic x coords={$H$=4, $H$=8, $H$=16},
    xtick=data,
    ymin=58,ymax=82,
    nodes near coords={\scriptsize\pgfmathprintnumber\pgfplotspointmeta},
    ]
\addplot+[error bars/.cd, y dir=both,y explicit] coordinates {
($H$=4,72.21) +- (0.0, 1.16)
($H$=8,72.56) +- (0.0, 1.23)
($H$=16,72.33) +- (0.0, 1.12) 
};
\addplot+[error bars/.cd, y dir=both,y explicit] coordinates {
($H$=4,76.54) +- (0.0, 2.02) 
($H$=8,76.43) +- (0.0, 1.89)
($H$=16,76.29) +- (0.0, 2.06)
};
\addplot+[error bars/.cd, y dir=both,y explicit] coordinates {
($H$=4,61.81) +- (0.0, 1.52)
($H$=8,61.79) +- (0.0, 1.55)
($H$=16,62.10) +- (0.0, 1.58)
};
\legend{\textsc{Cora},\textsc{Pubmed},\textsc{Citeseer}}
\end{axis}
\end{tikzpicture}
}
\\
\centering
\resizebox{6cm}{!}{
\begin{tikzpicture}
\begin{axis}[
    ybar,
    enlarge x limits=0.25,
    legend style={at={(0.5,1)},
                anchor=north,legend columns=3},
    ylabel={Accuracy},
    xlabel={Transductive},
    symbolic x coords={$T$=16, $T$=32, $T$=64, $T$=128},
    xtick=data,
    ymin=68,ymax=85,
    nodes near coords={\scriptsize\pgfmathprintnumber\pgfplotspointmeta},
    ]
\addplot+[error bars/.cd, y dir=both,y explicit] coordinates {
($T$=16,80.76) +- (0.0, 1.40)
($T$=32,81.09) +- (0.0, 1.39) 
($T$=64,80.92) +- (0.0, 1.29)
($T$=128,81.15)+- (0.0, 1.14)
};
\addplot+[error bars/.cd, y dir=both,y explicit] coordinates {
($T$=16,77.84) +- (0.0, 2.02)
($T$=32,77.99) +- (0.0, 1.78) 
($T$=64,77.99) +- (0.0, 2.08)
($T$=128,78.13)+- (0.0, 2.03)
};
\addplot+[error bars/.cd, y dir=both,y explicit] coordinates {
($T$=16,70.67) +- (0.0, 1.21)
($T$=32,70.36) +- (0.0, 1.16) 
($T$=64,70.49) +- (0.0, 1.28)
($T$=128,70.24)+- (0.0, 1.21)
};
\legend{\textsc{Cora},\textsc{Pubmed}, \textsc{Citeseer}}
\end{axis}
\end{tikzpicture}
}
\caption{Effects of the number $T$ of random walks, the number $K$ of attention layers and the number $H$ of attention heads on the validation sets. We fixed the same value for one hyper-parameter and tune other hyper-parameters for all 10 data splits of each dataset.}
\label{fig:EffectsLH}
\end{figure}
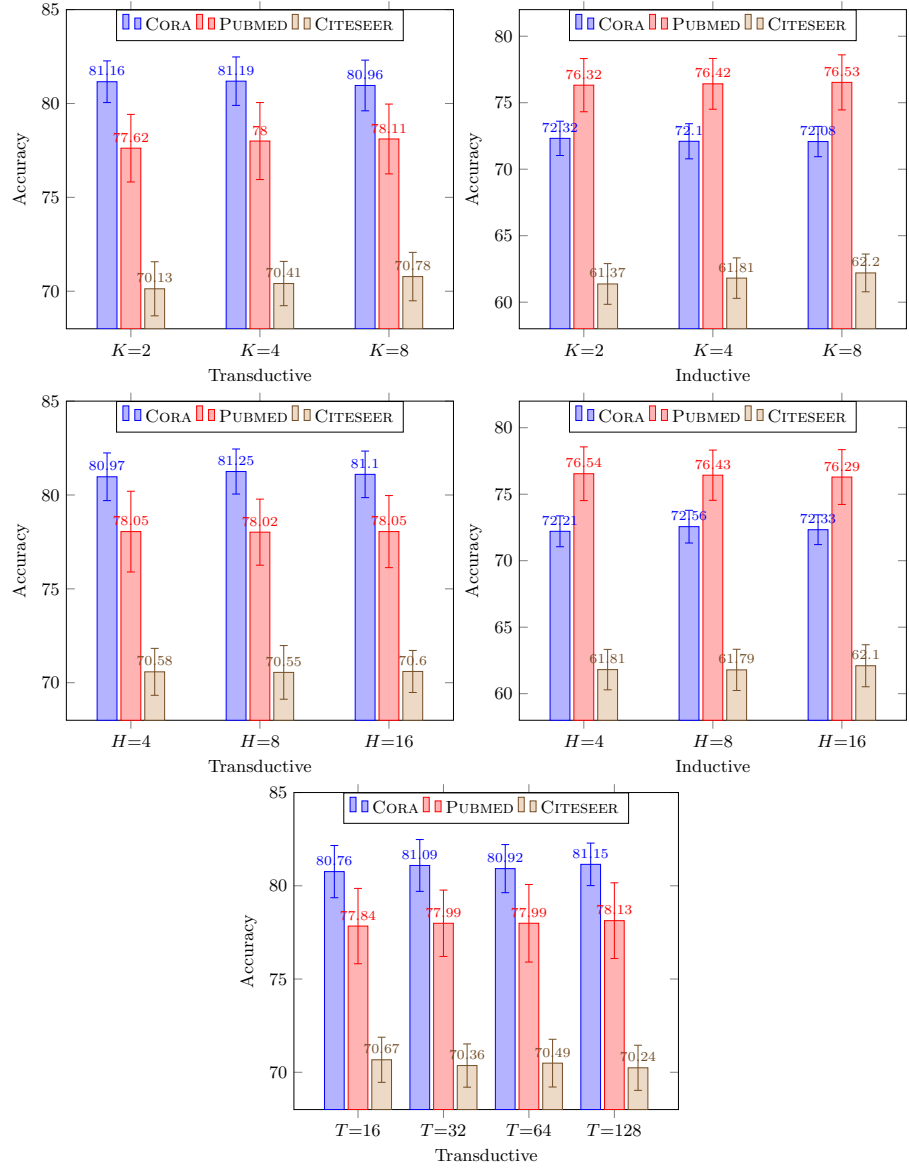

We compute and report our ablation results on the validation sets in the transductive setting over two factors in Table \ref{tab:expresultsDev}.
There is a decrease in the accuracy results when not using the feed-forward sub-layer, but we do not see a significant difference between with and without using this sub-layer (at $p <0.05$ using the two-tailed {paired t-test}). 
More importantly, without the multi-head self-attention sub-layer, the results degrade by more than 3.2\% on all three datasets, showing the merit of this self-attention sub-layer in learning the plausible node embeddings. 
Note that similar findings also occur in the inductive setting.

\subsection{Effects of hyper-parameters}

We investigate the effects of hyper-parameters on the \textsc{Cora}, \textsc{Pubmed}, and \textsc{Citeseer} validation sets of 10 data splits in Figure \ref{fig:EffectsLH}, when we use the same value for one hyper-parameter and then tune other hyper-parameters for all 10 data splits of each dataset.
Regarding the transductive setting, we see that the high accuracies can be generally obtained when using $T=128$ on \textsc{Cora} and \textsc{Pubmed}, and $T=16$ on \textsc{Citeseer}.
This is probably because \textsc{Citeseer} are more sparse than \textsc{Cora} and \textsc{Pubmed}, especially the average number of neighbors per node on \textsc{Cora} and \textsc{Pubmed} are 2.0 and 2.2 respectively, while it is just 1.4 on \textsc{Citeseer}.
This is also the reason why we set $T=128$ on \textsc{Cora} and \textsc{Pubmed}, and $T=16$ on \textsc{Citeseer} during training in the inductive setting.
Besides, regarding the number $K$ of attention layers for both the transductive and inductive settings, using a small $K$ produces better results on \textsc{Cora}. At the same time, there is an accuracy increase on \textsc{Pubmed} and \textsc{Citeseer} along with increasing $K$. 
Regarding the number $H$ of attention heads, we achieve higher accuracies when using $H=8$ on \textsc{Cora} in both the settings.
Besides, there is not much difference in varying $H$ on \textsc{Pubmed} and \textsc{Citeseer} in the transductive setting.
But in the inductive setting, using $H=4$ gives high scores on \textsc{Pubmed}, while the high scores on \textsc{Citeseer} are obtained by setting $H=16$.

\section{Conclusion}
\label{sec:conclusion}

We introduce a novel unsupervised embedding model SANNE to leverage from the random walks to induce the transformer self-attention network to learn node embeddings.
SANNE aims to infer plausible embeddings not only for present nodes but also for new nodes.
Experimental results show that our SANNE obtains the state-of-the-art results on \textsc{Cora}, \textsc{Pubmed}, and \textsc{Citeseer} in both the transductive and inductive settings.
Our code is available at: \url{https://github.com/daiquocnguyen/Walk-Transformer}.

\bibliographystyle{splncs04}
\bibliography{references}

\end{document}